\documentclass[unnumsec,webpdf,contemporary,large]{oup-authoring-template}%

\usepackage{booktabs}
\usepackage{longtable}

\usepackage{multirow}
\usepackage{tabularx}

\usepackage{tokcycle}

\newcommand\bverb[1][]{\begingroup\ifx\relax#1\relax
    \def\tmpA{\allowbreak}\let\tmpB=\empty 
      \setlength\spaceskip{1.5ex plus 1ex minus .5ex}
    \else\def\tmpA{}\let\tmpB=#1 
      \setlength\spaceskip{1.5ex plus 1ex minus .5ex}\fi
  \catcode`\%=12 \bverbaux}

\newcommand\bverbaux[2][ ]{%
   \ttfamily\tokencycle
   {\tctestifx{\tmpB##1}{\addcytoks{\allowbreak}}{%
    \addcytoks[1]{\tmpA}\addcytoks[1]{\string##1}%
    \addcytoks{\nobreak\hspace{0pt plus 5pt minus .5pt}}}}
   {\addcytoks[1]{\string{}\processtoks{##1}\addcytoks[1]{\string}}}
   {\tctestifx{\tmpB##1}{\addcytoks{\allowbreak}}
     {\tctestifnum{\cstest{##1}=11} 
      {\tcpeek\zzz\tctestifcatnx\zzz A{\tcpush{\space}}{}}{}%
      \tcpush{\string##1}}}
   {\addcytoks{\allowbreak#1}}#2\endtokencycle\endgroup}
\def\cstest#1{\expandafter\cstestaux\string#1.\relax}
\def\cstestaux#1#2#3\relax{\the\catcode`#2 }

\graphicspath{{Fig/}}

\theoremstyle{thmstyleone}%
\theoremstyle{thmstyletwo}%
\theoremstyle{thmstylethree}%

\begin{document}

\journaltitle{Journal Title Here}
\DOI{DOI HERE}
\copyrightyear{2022}
\pubyear{2019}
\access{Advance Access Publication Date: Day Month Year}
\appnotes{Paper}

\firstpage{1}


\title[Project Riley]{Project Riley: Multimodal Multi-Agent LLM Collaboration with Emotional Reasoning and Voting}

\author[1]{Ana Rita Ortigoso \ORCID{0009-0001-7529-5857}}
\author[1]{Gabriel Vieira \ORCID{0009-0000-2300-8441}}
\author[1]{Daniel Fuentes \ORCID{0000-0001-9726-1087}}
\author[1]{Luís Frazão \ORCID{0000-0003-2571-7940}}
\author[1]{Nuno Costa \ORCID{0000-0002-2353-369X}}
\author[1,$\ast$]{António Pereira \ORCID{0000-0001-5062-1241}}

\authormark{Ortigoso et al.}

\address[1]{\orgdiv{Computer Science and Communication Research Centre}, \orgname{Polytechnic University of Leiria}, \orgaddress{\street{Morro do Lena – Alto do Vieiro}, \postcode{2411-901}, \state{Leiria}, \country{Portugal}}}

\corresp[$\ast$]{Corresponding author. \href{email:apereira@ipleiria.pt}{apereira@ipleiria.pt}}

\received{Date}{0}{Year}
\revised{Date}{0}{Year}
\accepted{Date}{0}{Year}



\abstract{This paper presents Project Riley, a novel multimodal and multi-model conversational AI architecture oriented towards the simulation of reasoning influenced by emotional states. Drawing inspiration from Pixar’s Inside Out, the system comprises five distinct emotional agents—Joy, Sadness, Fear, Anger, and Disgust—that engage in structured multi-round dialogues to generate, criticise, and iteratively refine responses. A final reasoning mechanism synthesises the contributions of these agents into a coherent output that either reflects the dominant emotion or integrates multiple perspectives. The architecture incorporates both textual and visual large language models (LLMs), alongside advanced reasoning and self-refinement processes. A functional prototype was deployed locally in an offline environment, optimised for emotional expressiveness and computational efficiency. From this initial prototype, another one emerged, called Armando, which was developed for use in emergency contexts, delivering emotionally calibrated and factually accurate information through the integration of Retrieval-Augmented Generation (RAG) and cumulative context tracking. The Project Riley prototype was evaluated through user testing, in which participants interacted with the chatbot and completed a structured questionnaire assessing three dimensions: Emotional Appropriateness, Clarity and Utility, and Naturalness and Human-likeness. The results indicate strong performance in structured scenarios, particularly with respect to emotional alignment and communicative clarity.}


\maketitle

\section{Introduction}
\label{sc:introduction}

The `Inside Out' saga \cite{Docter2015, Mann2024}, produced by Pixar Animation Studios, provides a symbolic narrative depiction of the perceptions and emotional interactions inherent in human experience.

Throughout the development of the saga's films, the creators aimed not only to produce artistic entertainment but also to translate fundamental psychological concepts into a language accessible to the general public. Consequently, the films benefited from consultation with experts in psychology and neuroscience, including Dacher Keltner, Paul Ekman, and Lisa Damour \cite{Brice2015, Barber2024}.

The saga narrates the life of a girl named Riley, with the use of autonomous characters representing distinct emotions serving to personify her emotional experiences. The first film (2015) \cite{Docter2015}, is set during Riley's childhood and focuses on five primary emotions: Joy, Sadness, Fear, Anger, and Disgust. The sequel (2024) \cite{Mann2024} portrays Riley's adolescence and the complexities typical of this developmental stage, introducing additional emotions: Shame; Envy; Boredom; and Anxiety. These emotions interact dynamically, guiding Riley's behavioural responses to life changes and challenges.

Inspired by the films, this paper explores the development of a conversational generative Artificial Intelligence (AI) solution that employs an approach analogous to the emotional interplay that guides Riley's behaviour. The objective is to generate emotionally informed responses, similar to those produced by human beings who are emotionally aware.

Therefore, this paper presents the architecture of the generative AI conversational solution, Project Riley. This system is characterised by its emotional awareness, multi model functionality, and the implementation of sophisticated self-refining and voting logic. Project Riley integrates multimodal information, text and images, through the coordinated use of multiple Large Language Models (LLM), each representing a distinct emotion. It processes user input using both standard and multimodal LLM for textual and visual analysis, respectively, and orchestrate multi-round emotional dialogues. The final output results from a structured voting and synthesis mechanism that uses an advanced reasoning model to  combine emotionally differentiated responses into a coherent answer. This architecture demonstrates the potential of GenAI to combine the generative capabilities of neural models with symbolic reasoning structures, such as agent-specific roles, conversational history tracking, voting with justification, and formally segmented outputs, thereby enabling interpretation and decision-making across heterogeneous data sources.

A emergency-oriented variant of the system, named Armando, is also presented, which architecture was adaptedto real-time, high-stakes environments with a focus on trust, relevance, and emotional regulation.

To the best of our knowledge, this is the first framework that leverages generative AI to orchestrate structured affective reasoning through independent emotional agents, introducing a novel paradigm for integrating generative emotional modelling into conversational systems. Furthermore, no comparable solution has been identified in the literature that adapts such an architecture to emergency response contexts as implemented in Armando, where emotional regulation, informational accuracy, and user trust are prioritised under high-stakes conditions.

\section{Related Work}

\subsection{Emotion Recognition and Emotional Intelligence}

Recent literature on emotionally intelligent systems primarily addresses the integration of emotional awareness into AI models, particularly LLM. Ratican et al. \cite{Ratican2023} introduced the 6DE conceptual model, featuring six emotional dimensions — arousal, valence, dominance, agency, fidelity, and novelty — to provide a comprehensive framework for analysing human emotions in AI systems. This approach aims to enhance empathy and contextual relevance, specifically targeting applications in education, mental health, and assisted care. Their suggested integration methods include emotion-guided prompting, annotated datasets, user feedback, emotional database integration, and emotional quality refinement through post-processing.

Kang et al. \cite{Kang2024} similarly focused on emotional integration, albeit within a social robotic context, employing the Pleasure-Arousal-Dominance (PAD) emotional model. They developed the SoR-ReAct agent for the Nadine robot platform, which combines GPT-4 with personalised episodic memory, enabling emotional simulation and personalised interactions based on historical data. Unlike Ratican et al., Kang et al. directly incorporated episodic memory retrieval facilitated by OpenAI's embeddings, enhancing the robot’s interactive empathy capabilities through multimodal perception.

In contrast, Liu et al. \cite{Liu2025} addressed emotional intelligence specifically within negotiation scenarios, developing EQ-Negotiator, which integrates emotion recognition (RoBERTa, GPT-2, DeBERTa) with strategic game-theory-based reasoning. Their work distinctly emphasised balancing emotional empathy and assertiveness, using Hidden Markov Models for dynamic emotional adaptations. Notably, their approach outperformed GPT-4 in terms of emotional realism and negotiation efficacy.

Rasool et al. \cite{Rasool2024} and Liu et al. \cite{Liu2024} extended emotional intelligence into psychotherapy and multimodal dialogue, respectively. Rasool et al. employed emotional lexicons (NRC Emotion Lexicon, SentiWordNet) combined with hierarchical segmentation and attention mechanisms to enhance empathetic responses, despite noting a trade-off between empathy and coherence. Liu et al., conversely, integrated emotional and visual modalities using Emotional Retrieval Module(ERM), Response Emotion Prediction (REP), and Emotion-Enhanced Response Generation (EERG) models. Their multimodal ELMD system demonstrated superior emotional and contextual responsiveness compared to traditional dialogue systems.

Chen et al. \cite{Chen2024}, Motwani et al. \cite{Motwani2024}, and Yang et al. \cite{Yang2025} further highlighted emotional considerations, albeit within specialised frameworks. Chen et al.'s RECONCILE framework collaboratively utilised multiple LLM agents to generate emotionally and contextually relevant responses, leveraging weighted voting based on reasoning justification. Motwani et al. proposed a sequential training method employing specialised agents in reasoning and verification tasks, indirectly enhancing emotional intelligence through robust error correction. Yang et al.'s SupportlyChat directly focused on therapeutic contexts using Cognitive Behavioural Therapy (CBT), combining RoBERTa and sentiment analysis to enhance emotional awareness and professional interaction in mental health support scenarios.

Finally, Brun et al. \cite{Brun2025} evaluated emotional sensitivity in technical support chatbots, specifically employing VADER sentiment analysis within ChatGPT-3.5. Their findings indicated emotional sensitivity significantly improved user-perceived competence, empathy, trust, and continued use intention, although without measurably affecting user emotional states.

\subsection{Multi-Agent Debate and Collaborative Reasoning}

Research exploring multi-agent systems frequently emphasises collaborative reasoning and decision-making through debate mechanisms. Zhao et al. \cite{Zhao2024} tested electoral methods from social choice theory (e.g., Borda, IRV, Ranked Pairs) in collective decision-making scenarios involving multiple LLM agents, highlighting notable accuracy improvements in benchmarks like MMLU and ARC. They demonstrated that even minimal agent ensembles significantly enhance performance and recommended task-specific tailoring of decision methods.

Yang et al. \cite{Yang2025a} combined adversarial debate, weighted voting, and internal self-correction within a multi-agent framework to mitigate hallucinations and enhance reasoning efficiency. Their methodology involved iterative debate, error documentation, and dynamic weighting, outperforming baseline methods in accuracy and response speed on tasks like GSM8K and MMLU.

Similarly, Xu et al. \cite{Xu2023} proposed a peer-review inspired collaborative method. Agents independently generated solutions, reviewed peers' contributions, and refined responses accordingly. This explicit feedback and diverse model integration consistently improved performance across multiple datasets (GSM8K, StrategyQA, ARC-c).

Collectively, these studies underline the efficacy of structured multi-agent frameworks, emphasising the critical role of model diversity, debate dynamics, and systematic feedback in enhancing collaborative reasoning and overall system robustness.

\vspace{1em}
\noindent
In contrast to existing approaches, the architecture proposed in this work distinguishes itself by employing a structured multi-agent framework in which each agent represents a distinct emotion and participates in iterative emotional reasoning. While prior studies tend to integrate emotion either through static embeddings, sentiment lexicons, or personalised memory modules \cite{Ratican2023, Kang2024, Rasool2024}, this architecture fosters dynamic emotional dialogue and collective deliberation. Additionally, the integration of symbolic reasoning mechanisms—such as voting and justification—enhances explainability, a feature largely absent in previous neural architectures \cite{Chen2024, Motwani2024}. The system is also extensible to different emotional models and has been adapted for high-stakes domains, such as emergency response, through the Armando variant. To the best of our knowledge, no previous work combines emotional simulation, multimodal reasoning, and factual grounding within a unified architecture.

\section{Proposed Architecture}

This paper presents an innovative conversational AI system architecture that processes user queries through distinct emotional lenses, specifically a set of five basic emotions: Anger, Joy, Sadness, Fear, and Disgust. The system follows the emotional framework portrayed in the first Inside Out film, which was inspired by Paul Ekman’s theory of universal emotions \cite{Ekman1971}. Although Ekman initially identified six basic emotions—Happiness, Anger, Sadness, Disgust, Surprise, and Fear—and later expanded the list to include emotions such as Contempt and Enjoyment \cite{Ekman}, the film deliberately reduced the set to five for narrative clarity \cite{Cannon2016, Roper2015}. In this work, the same five emotions were retained to preserve the analogy with the film, thereby fostering greater familiarity and intuitive use—particularly among users who have seen the film. However, it is important to emphasise that the Project Riley architecture is not limited to this configuration: it is capable of simulating a broader range of emotions, depending on the requirements of the application and the computational resources available.

The system can be easily adapted to reflect alternative emotional models, allowing developers to incorporate different sets of emotions based on specific application needs or cultural relevance. Furthermore, the proposed architecture can be aligned with other psychological theories, such as Plutchik’s Wheel of Emotions \cite{Plutchik1982}, Carroll Izard’s Differential Emotions Theory \cite{Izard1977}, Jaak Panksepp’s Affective Neuroscience framework \cite{Panksepp1982}, or dimensional approaches like Russell’s Circumplex Model \cite{Russell1980} and the Pleasure-Arousal-Dominance (PAD) model \cite{Mehrabian1980}. Such adaptations would require architectural modifications and prompt reengineering to ensure consistency with the theoretical assumptions of the chosen model.

\begin{figure*}[!htbp]
    \includegraphics[width=\textwidth]{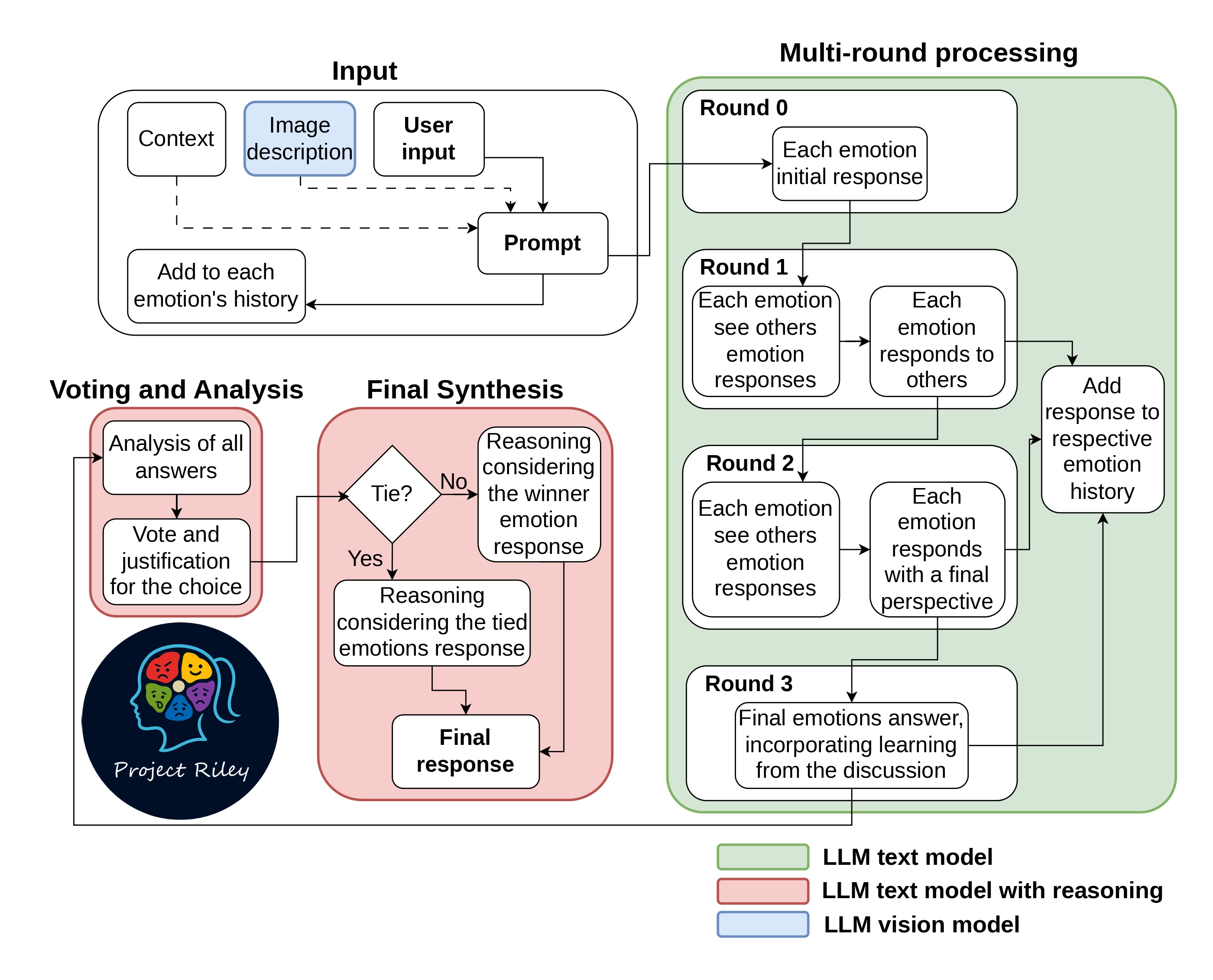}
    \caption{Proposed architecture}
    \label{fig:proposed_arch}
\end{figure*}

The architecture, as depicted in \autoref{fig:proposed_arch}, comprises four sequential phases: Input; Multi-round Processing; Voting and Analysis; and Final Synthesis.

Upon initialisation, the system establishes separate conversation histories for each emotional agent.

In the Input phase, users can provide contextual information and images prior to interaction. These images are described by a vision-based LLM (vision LLM), enriching the conversational context through visual data.

The processing workflow begins with the user input, which is injected as emotion-specific context into the conversation history of each agent.

Subsequently, the Multi-round Processing phase is initiated, employing a textual LLM. Each emotional agent independently generates an initial response, concurrently processed and reflecting the distinct emotional viewpoint (Round 0). These initial responses are documented and serve as the basis for subsequent interactions.

In the first discussion phase (Round 1), emotional agents review and respond to the initial outputs from the other emotions, maintaining emotional authenticity while engaging in dynamic interaction. This process progresses to a second discussion round (Round 2), in which each emotional agent synthesises a refined perspective based on insights gained from the prior exchange.

The third round (Round 3) involves the emotional agents reassessing the original user query, delivering evolved and finalised answers influenced by the preceding discussions. This iterative approach enriches the emotional discourse and is systematically captured within individual emotion-specific conversation histories.

In the Voting and Analysis phase, each emotional agent transitions to a reasoning model while preserving its distinct emotional identity. These reasoning models perform a critical evaluation of the candidate responses, casting votes accompanied by concise justifications. The system then aggregates these votes to determine a consensus or identify a tie.

The Final Synthesis also utilises a reasoning model to integrate the emotional perspectives indicated by the voting outcomes. If a single emotional perspective predominates, the final response primarily reflects insights from the winning emotion. In the event of a tie, the response is synthesised from multiple emotional perspectives, ensuring balance. The final output is clearly segmented into REASONING (analytical assessment), THOUGHTS (representing Riley’s internal cognitive processes), and FINAL ANSWER (a balanced and consumable synthesis for the user).

At the end of the process, all interactions, dialogues, discussions, and relevant information should be made available to the user.

\section{Project Riley chatbot prototype} \label{sec:riley_prototype}

In accordance with the proposed architecture, a general-purpose chatbot was implemented based entirely on the Project Riley architecture. To enable local deployment of LLM, the Ollama framework was utilised \cite{Ollama2025}. The hardware specifications of the machine used for this implementation are detailed in Table \ref{tab:computer_specs}.

\begin{table}[!htbp]
\caption{Computer Specifications}
\centering
\begin{tabular}{|p{2cm}|p{5.5cm}|}
\hline
\textbf{Component} & \textbf{Specification} \\
\hline
Processor & Intel Core i3-9100 (4 cores, 4 threads, 3.6~GHz base clock) \\
\hline
Memory & 8~GB DDR4 RAM \\
\hline
Graphics Card & NVIDIA GeForce RTX 3070, 8~GB GDDR6 VRAM \\
\hline
\end{tabular}
\label{tab:computer_specs}
\end{table}

Given the hardware constraints, particularly regarding Graphics Processing Unit (GPU) Video Random Access Memory (VRAM) capacity, considerable attention was directed towards selecting models that offer a balanced trade-off between performance and model size. Furthermore, when computationally feasible, we prioritised the use of abliterated models. Abliteration is a technique that alters the weights of a model to bypass default safety alignment mechanisms typically introduced during fine-tuning \cite{Labonne2024}. These mechanisms, while intended to ensure safe output, often over-filter emotionally charged or controversial prompts, leading to overly polite, neutral, or evasive responses.

Through empirical testing, we observed that abliterated models consistently produced more genuine and emotionally resonant outputs, better aligned with the behavioural profiles expected from each emotional agent. This enhanced expressive freedom was deemed essential for the fidelity of the multi-agent emotional simulation at the core of Project Riley.

Consequently, the chosen text-based LLM model was \bverb{huihui_ai/llama3.2-abliterate:3b} \cite{Meta}, which has 3.61 billion parameters and a model size of 2.2 GB. For image description tasks, the selected LLM was \bverb{gemma3:4b} \cite{Google}, a model capable of processing both text and images, comprising 4.3 billion parameters and a model size of 3.3 GB. For advanced reasoning tasks, the \bverb{huihui_ai/deepseek-r1-abliterated:8b} \cite{Deepseek} model was utilised, containing 8.03 billion parameters and a model size of 4.9 GB.

This selection of models was designed to maximise the likelihood that they remain loaded in GPU VRAM, thereby reducing the need for frequent model switching and minimising latency between the initial prompt and the final response. Nonetheless, when an image description is requested, one model must be unloaded to load the reasoning model, which introduces additional processing delay.

All interactions and computational processes within Project Riley are meticulously logged to ensure transparency. Users are provided with the option to download comprehensive conversation logs, which include messages, model parameters, and contextual information.

\subsection{Prototype Evaluation} \label{subsec:prototype_evaluation}

The developed chatbot was evaluated through user testing with 17 participants, wherein they interacted with the system and provided feedback via a questionnaire. Participants were instructed to address five themes: Job Loss or Unemployment; Breakups or Friendship Loss; Difficult Personal Decisions; Anxiety in Academic or Professional Contexts; and Family or Intergenerational Conflicts. As detailed in \autoref{tab:questions}, the questions presented in the questionnaire were categorised into three main dimensions: Emotional Appropriateness, Clarity and Utility, and Naturalness and Humanisation. Most questions utilised a 5-point Likert scale, where 1 indicated "Strongly disagree" and 5 indicated "Strongly agree". An exception was the question "Which emotion do you predominantly identify in the final response?", which allowed participants to provide an open-text answer.

\begin{table*}[!htbp]
\centering
\caption{Questions}
\begin{tabularx}{\textwidth}{|l|X|}
\hline
\textbf{Category} & \textbf{Question description} \\
\hline
\multirow{4}{*}{Emotional Appropriateness} 
& - Did the final response convey empathy? \\
& - Did the final response seem emotionally appropriate to the context of your question? \\
& - Do you believe the answer(s) of the emotion(s) with the most votes were the most appropriate? \\
& - Which emotion do you predominantly identify in the final response? \\
\hline

\multirow{4}{*}{Clarity and Utility} 
& - Was the final response clear and understandable? \\
& - Was the final response consistent with your original question? \\
& - Was the final response useful or did it help you reflect on your question? \\
& - Did the visualisation of the process leading to the final response help you reflect on your question? \\
\hline

\multirow{1}{*}{Naturalness and Human-likeness} 
& - Did the response feel as though it was written by a human? \\
\hline
\end{tabularx}
\label{tab:questions}
\end{table*}

\begin{figure*}[!htbp]
\centering
\includegraphics[width=0.65\linewidth]{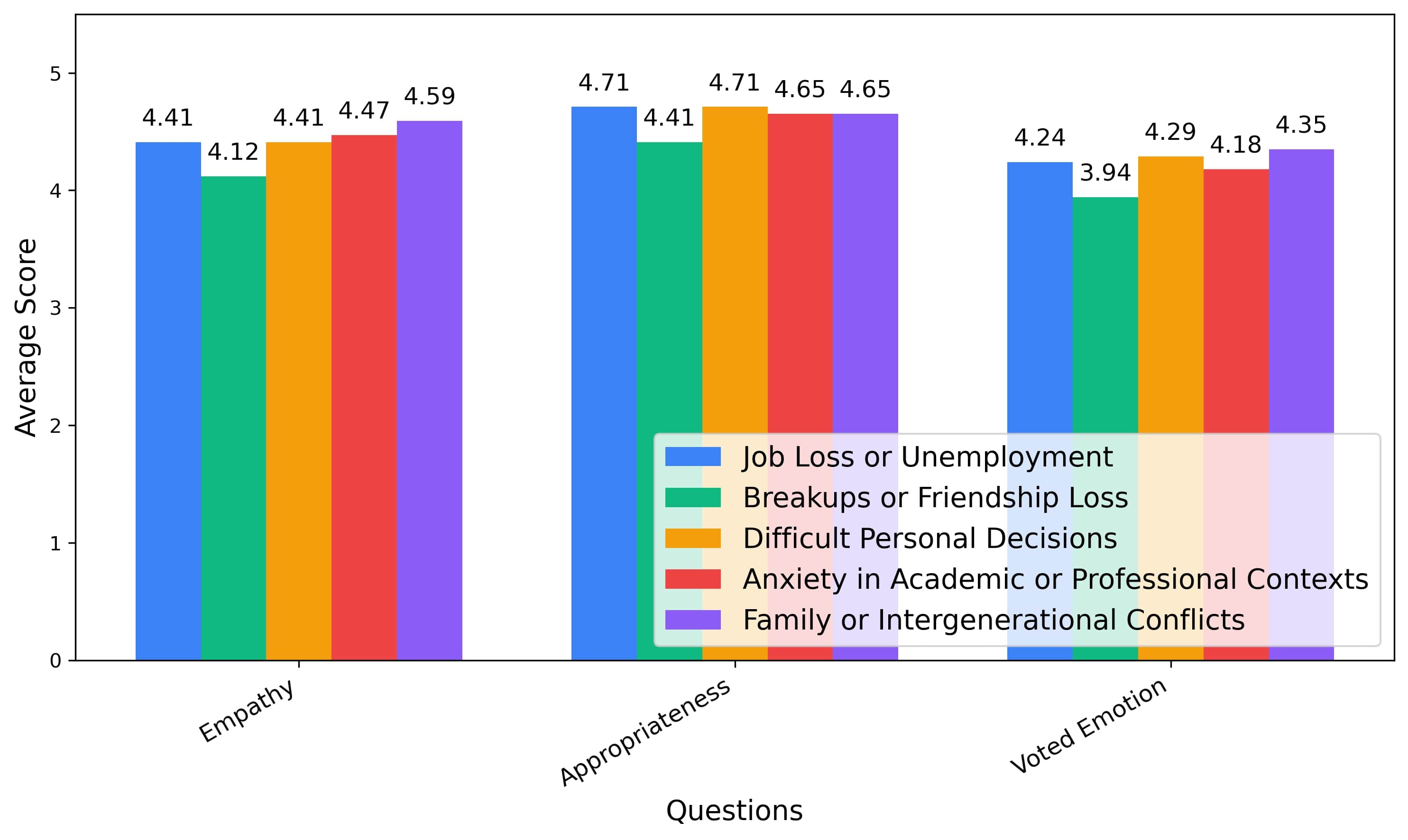}
\caption{Emotional Appropriateness questions average scores across sections}
\label{fig:Emotional Appropriateness}
\end{figure*}

Average scores for Emotional Appropriateness across themes are shown in \autoref{fig:Emotional Appropriateness}. For the question "Did the final response convey empathy?", the highest average was recorded in Family or Intergenerational Conflicts (4.59, mode=5), suggesting strong empathetic resonance. Conversely, Breakups or Friendship Loss had the lowest score (4.12, mode=4), indicating difficulties in effectively conveying empathy in interpersonal loss scenarios.

Concerning the question “Did the final response seem emotionally appropriate to the context of your question?”, highest averages (4.71, mode=5) appeared in Job Loss or Unemployment and Difficult Personal Decisions, signifying well-calibrated emotional tone in professional and personal contexts. Breakups or Friendship Loss again showed the lowest average (4.41), reinforcing the difficulty in addressing relational loss situations adequately.

For the question \textit{Do you believe the answer(s) of the emotion(s) with the most votes were the most appropriate?}, the alignment of system-selected emotions with user expectations was strongest in Family or Intergenerational Conflicts (4.35). Breakups or Friendship Loss remained the lowest scoring (3.94), consistently reflecting lower emotional alignment.

\begin{figure*}[!htbp]
\centering
\includegraphics[width=0.65\linewidth]{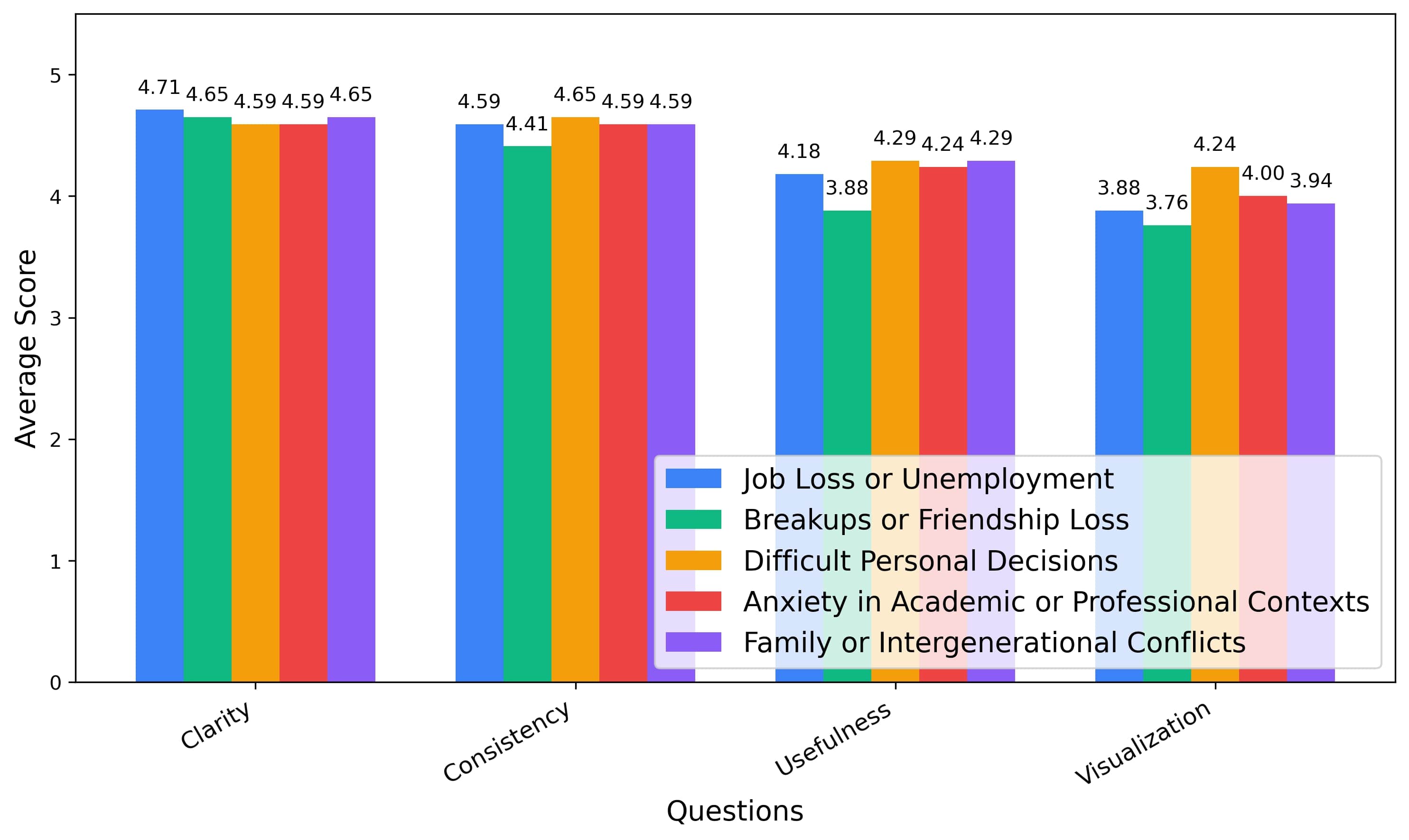}
\caption{Clarity and Utility questions average scores across sections}
\label{fig:clarity_utility}
\end{figure*}

Clarity and Utility evaluations, depicted in \autoref{fig:clarity_utility}, highlighted high clarity across themes, particularly for the question \textit{Was the final response clear and understandable?}, with Job Loss or Unemployment receiving the highest score (4.71, mode=5). The lowest consistency with the original question, \textit{Was the final response consistent with your original question?}, was observed for Breakups or Friendship Loss (4.41), while Difficult Personal Decisions attained the highest consistency (4.65). Regarding the question \textit{Was the final response useful or did it help you reflect on your question?}, scores varied, with Breakups or Friendship Loss lowest (3.88), and Difficult Personal Decisions and Family or Intergenerational Conflicts highest (4.29). Visualisation support, assessed through the question \textit{Did the visualisation of the process leading to the final response help you reflect on your question?}, received the lowest overall scores, ranging from 3.76 to 4.24, indicating limited perceived benefit.

\begin{figure*}[!htbp]
\centering
\includegraphics[width=0.65\linewidth]{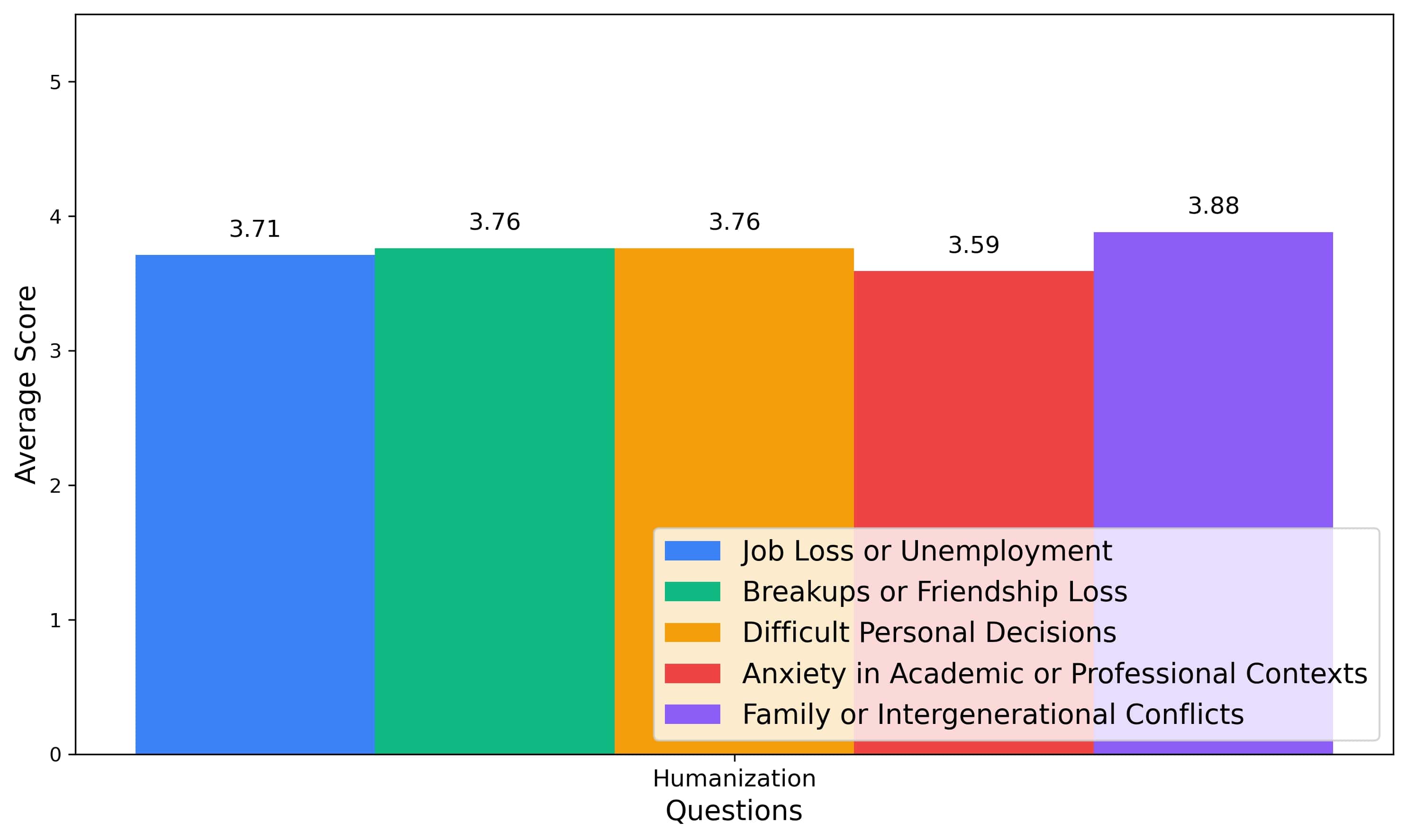}
\caption{Humanization question average scores across sections}
\label{fig:humanization}
\end{figure*}

The perception of human-likeness (\autoref{fig:humanization}), measured by the question \textit{Did the response feel as though it was written by a human?}, indicated moderate human-like communication, with the highest score in Family or Intergenerational Conflicts (3.88). Anxiety in Academic or Professional Contexts (3.59) again had the lowest human-like perception, highlighting continued challenges in generating naturally nuanced responses to relational losses.

Analysis of responses to the question \textit{Which emotion do you predominantly identify in the final response?} (\autoref{tab:predominant_emotions}) reveals significant variability in emotional perception. In the Anxiety in Academic or Professional Contexts theme, Joy was most frequently identified (7 mentions), possibly reflecting a reassuring approach adopted by the system. Conversely, Sadness dominated responses related to Breakups or Friendship Loss (5 mentions), indicating the appropriateness of a more  reflective emotional tone in relational loss scenarios. Difficult Personal Decisions equally elicited Fear and Joy (5 mentions each), suggesting responses were balanced between cautionary and optimistic perspectives. Family or Intergenerational Conflicts predominantly evoked Joy (6 mentions), again highlighting positivity in familial contexts. Lastly, Job Loss or Unemployment mainly elicited Fear (5 mentions), aligning well with the typically uncertain and stressful nature of employment-related concerns.

\begin{table*}[!htbp]
\centering

\caption{Predominant emotions per section}
\begin{tabular}{p{7cm}p{9.5cm}}
\toprule
Section & Predominant Emotions \\
\midrule
\textbf{Anxiety in Academic or Professional Contexts} & Joy (7), Sadness (3), Fear (2), Anger (1), Disgust (1), Empathy (1), Enthusiasm (1), Hope (1) \\
\textbf{Breakups or Friendship Loss} & Sadness (5), Joy (4), Disgust (3), Empathy (2), Fear (2), Neutral (1) \\
\textbf{Difficult Personal Decisions} & Fear (5), Joy (5), Empathy (3), Sadness (3), Disgust (1) \\
\textbf{Family or Intergenerational Conflicts} & Joy (6), Empathy (3), Sadness (3), Fear (2), All (1), Confidence (1), Disgust (1) \\
\textbf{Job Loss or Unemployment }& Fear (5), Joy (4), Anger (3), Sadness (2), Confidence (1), Disgust (1), Empathy (1) \\
\bottomrule
\label{tab:predominant_emotions}
\end{tabular}
\end{table*}

\section{Armando: A emergency response chatbot using Project Riley}

Based on the Riley architecture, another prototype was developed, a chatbot named Armando that has been developed specifically to assist during emergency response scenarios. This AI assistant aims to provide citizens with emotionally-aware, human-like interactions, delivering responses grounded not only in officially validated information provided by authoritative sources but also crafted to mitigate panic and maintain user calmness during critical situations.

This approach seeks to mitigate issues such as those observed on 28 April 2025, during a widespread power outage across the Iberian Peninsula, when some individuals turned to ChatGPT for information. As the chatbot was not updated with official communications from public safety authorities, its use led to the dissemination of unreliable and inaccurate information, thereby increasing public confusion and anxiety \cite{SAPO242025}.

To ensure that Armando's responses accurately reflect official guidelines and authoritative information, a Retrieval-Augmented Generation (RAG) approach has been implemented. RAG enhances chatbot interactions by integrating external knowledge-context injection-during chatbot response generation. Specifically, a diverse set of authoritative sources—comprising official documents and information extracted from verified webpages—is indexed through semantic embeddings. These embeddings, generated using the \bverb{mxbai-embed-large} model in this case, provide a structured representation of the underlying content in a high-dimensional vector space. When a user submits a query, its semantic embedding is computed and compared against the indexed document embeddings. The content with the highest semantic similarity is then retrieved and used to ground the system’s response in accurate and contextually relevant information. When a user interacts with Armando, the chatbot retrieves relevant information associated with embeddings that closely match the query context, enriching its generated responses with verified external knowledge.

For effective and contextually relevant embedding retrieval, a cumulative context mechanism was developed. This mechanism continuously stores and updates key conversation topics, relevant keywords, the three most recent user questions, and a dynamically generated summary of the ongoing interaction. Such cumulative context ensures that retrieved embeddings remain consistently aligned with the evolving conversational context.

Given the specific requirements of this emergency response application, namely, reduced response times and lower computational overhead-targeted optimisations were implemen\-ted. The Final Synthesis and Voting and Analysis processes were simplified to enhance execution speed and reduce GPU VRAM usage. To this end, prompt modifications were introduced to emulate a reasoning logic using a conventional text-based LLM. Furthermore, as the THOUGHTS section was deemed non-essential in an emergency context and its exclusion contributed to faster response generation, it was omitted from the chatbot’s Final Synthesis.

\begin{figure}
    \centering
    \includegraphics[width=3.5in]{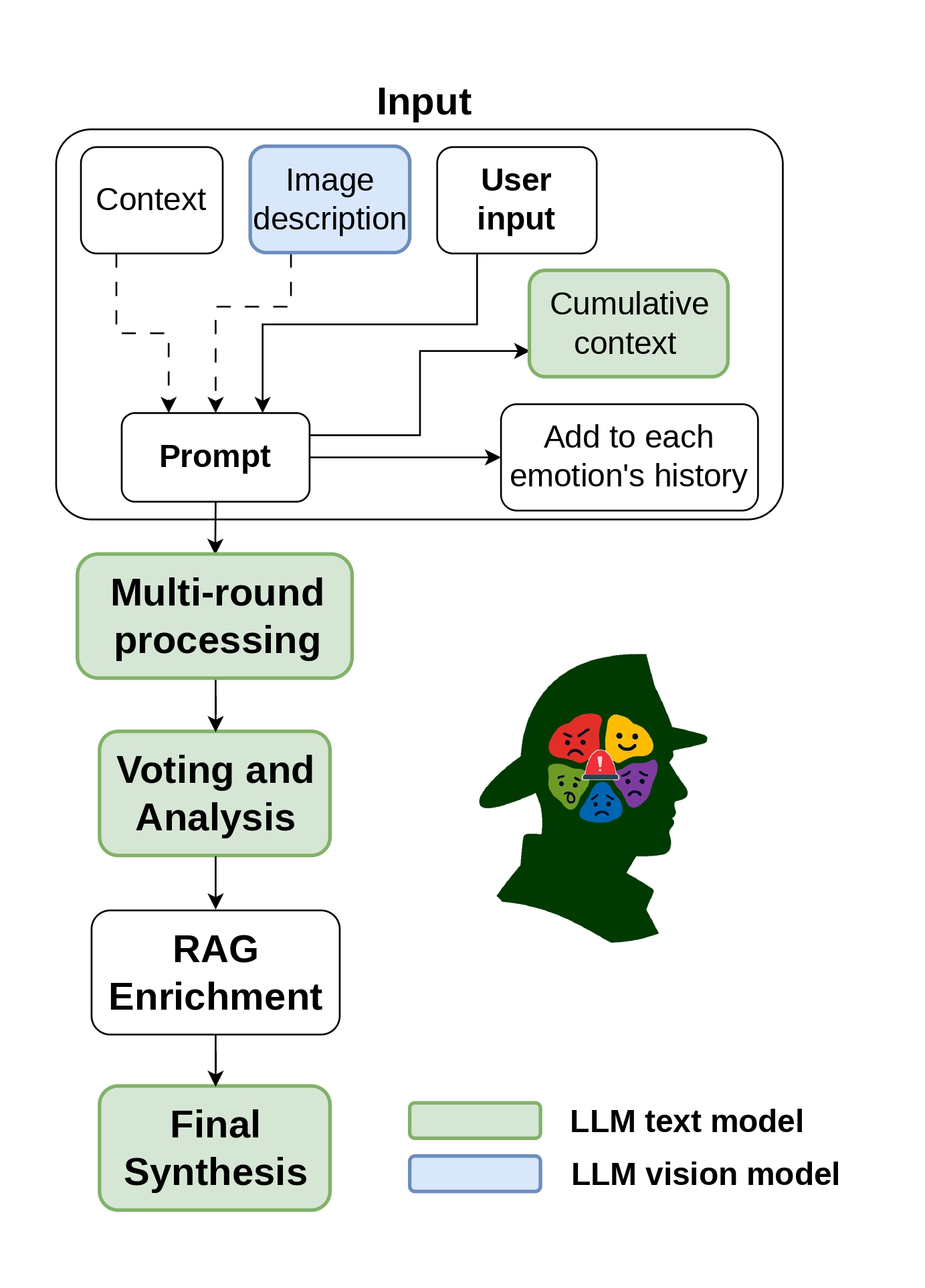}
    \caption{Armando architecture}
    \label{fig:armando_arch}
\end{figure}

These architectural modifications are illustrated in \autoref{fig:armando_arch}, which presents the Armando chatbot architecture derived from the original Riley architecture.

As illustrated in \autoref{fig:armando_arch}, RAG enrichment is applied immediately before the final synthesis stage, contrary to what might appear more intuitive, which would be to introduce it prior to the multi-round processing. Such an approach would provide each emotional agent with validated information from the outset, allowing the entire discussion to be grounded in authoritative content. However, initial development and empirical testing revealed that this strategy does not reliably ensure the factual accuracy of the final output, as critical information may become diluted or distorted during the emotional dialogue. Therefore, RAG enrichment was instead positioned just before the synthesis stage, ensuring that the final response maintains emotional balance and contextual sensitivity, while also conveying accurate and up-to-date information to the user.

The models used in this solution are the same that had been utilised in \autoref{sec:riley_prototype}.

\subsection{Response Comparison: RAG vs No RAG}

To assess the impact of RAG on the performance of the Armando chatbot, a series of prompts were issued in both RAG and non-RAG configurations. The RAG-enhanced responses were generated using five embedded documents: one describing the current emergency (including date, start time, incident status, type, summary, coordinates, and full address), another listing emergency contact numbers worldwide, and three others providing procedural guidance for earthquakes, fires, and floods.

The first prompt consisted solely of the user's query: "Where is the fire happening?". Responses obtained with and without RAG are presented in \autoref{tab:rag_where}.

\begin{table*}[!htbp]
\centering
\caption{Comparison of AI responses with and without RAG to the question: "Where is the fire happening?"}
\label{tab:rag_where}
\begin{tabular}{@{\extracolsep{\fill}} p{3cm} p{13cm}}
\toprule
\textbf{Response Type} & \textbf{Response Content} \\
\midrule

\textbf{Without RAG} & 
\textit{If you are in a safe location and hear the sound of breaking glass or a loud noise, do not panic – it may indicate that help is near.} \newline
• Check your surroundings for any signs of fire, such as flames, smoke, or heat. \newline
• Follow evacuation procedures outlined by local authorities, which should include clear routes to safety and emergency contact information. \newline
• Stay informed through trusted news sources or emergency alerts from local governments or fire departments. \newline
• Remember that fire department teams are trained to respond quickly and effectively – your priority is getting to a safe location while following their instructions. \\
\midrule

\textbf{With RAG} & 
\textit{The fire is currently happening on the third floor of a residential apartment building located at Rua de São Bento 112, 1200-820 Lisboa, Portugal (coordinates: 38.7169° N, 9.1399° W).} \newline
– Local fire services are on site, and civil protection authorities have been notified to ensure a coordinated response. \newline
– A safety perimeter has been established around the building, and at least two individuals have been evacuated due to the fire. \newline
– The building is partially evacuated, but residents are advised to remain cautious and follow evacuation instructions from emergency personnel. \newline
– To stay informed and receive updates on the situation, please tune into local news broadcasts or check official social media channels. \newline
– Recommended emergency instructions include: \newline
\quad • Sound the alarm and call emergency services – activate the nearest fire alarm and dial your local emergency number. \newline
\quad • Evacuate immediately – do not attempt to fight large fires. Use the nearest safe exit. \newline
\quad • Do not use lifts – use the stairs to avoid danger. \newline
\quad • Stay low to the ground – crawl below the smoke where the air is cleaner. \newline
– Emergency services are working diligently to resolve the situation. \newline
\textit{Please stay calm, follow these instructions, and prioritise your safety. Further updates will follow.} \\
\bottomrule
\end{tabular}
\end{table*}

Analysing \autoref{tab:rag_where}, the non-RAG model lacked specific contextual information regarding the ongoing incident, resulting in a generic response containing broad emergency guidance without specifying the fire's location. Conversely, the RAG-enhanced model delivered a precise answer, clearly identifying the fire's location as the third floor of a residential apartment at Rua de São Bento 112, Lisbon, Portugal, including coordinates (38.7169° N, 9.1399° W), evacuation status, and detailed safety instructions. This demonstrates the significant advantage of employing RAG for delivering contextually relevant and actionable information in emergency scenarios.

In the second prompt, the user asked: ``A fire is happening! What should I do?'' with the context ``It's difficult to breathe.''. The responses obtained are detailed in \autoref{tab:rag_breathing}.

\begin{table*}[!htbp]
\centering
\caption{Comparison of AI responses with and without RAG to the question: \textit{"A fire is happening! What should I do?"} with the context: \textit{"It's difficult to breathe."}}
\label{tab:rag_breathing}
\begin{tabular}{@{\extracolsep{\fill}} p{3cm} p{13cm}}
\toprule
\textbf{Response Type} & \textbf{Response Content} \\
\midrule

\textbf{Without RAG} & 
• \textbf{Stay calm}: When faced with a fire emergency, take a moment to assess your surroundings and prioritise your safety. Try not to panic, as this can cloud your judgment and lead to poor decision-making. \newline
• \textbf{Get out quickly}: If you're in the building, exit through the nearest door. Don't rely on elevators, as they may be unavailable during a fire. Move swiftly and stay low to the ground to avoid inhaling smoke and toxic fumes. \newline
• \textbf{Call for help}: Once outside, call 911 or your local emergency number immediately. Provide your location and any relevant information about the fire. \newline
• \textbf{Meet at a safe spot}: Designate a meeting point outside the building where you can reunite with family members or others who may be in danger. This will ensure everyone's safety and provide an opportunity to account for each other's whereabouts. \newline
• \textbf{Prioritise bravery, not bravado}: If you do enter the building to rescue someone or retrieve essential items, remember that bravery is about taking action, not showing off. Stay focused on your goal and prioritise your own safety above all else. \newline
• \textbf{Seek help from authorities}: Once you're safe outside, allow emergency responders to assess the situation and provide assistance. They will have the necessary equipment and training to handle the fire and ensure everyone's well-being. \newline
\textit{Remember, in a fire emergency, every second counts. Stay calm, follow these steps, and prioritise your safety above all else.} \\
\midrule

\textbf{With RAG} & 
• Sound the alarm and call emergency services – Activate the nearest fire alarm if available, and dial your local emergency number to report the fire. \newline
• Evacuate immediately – Do not attempt to fight large fires. Leave the building calmly but quickly, using the nearest safe exit. \newline
• Use a wet cloth to cover your nose and mouth while evacuating – This helps reduce smoke inhalation as you leave the building. \newline
• Check doors before opening them – Use the back of your hand to feel if a door is hot. If it's not, open the door cautiously – the fire may be on the other side. \newline
• Once outside, go to the designated assembly point and report your presence – Wait for further instructions from emergency personnel. \newline
• Stay low to the ground while evacuating through smoke-filled areas – It reduces your exposure to toxic fumes. \newline
\textit{Remember, getting out of a burning building safely is the top priority. Don't worry about saving belongings or pets; focus on your own survival first. Help will arrive soon, and you'll be able to re-evaluate what's most important once the situation has been controlled.} \\
\bottomrule
\end{tabular}
\end{table*}

From \autoref{tab:rag_breathing}, both RAG and non-RAG responses effectively addressed the user's situation by providing clear, practical evacuation and safety instructions suitable for scenarios involving breathing difficulties due to smoke inhalation.

Finally, the third prompt included the user's query: "Which number should I call?" with the context "I'm alone." and an accompanying image depicting a fire. The responses are summarised in \autoref{tab:rag_call_number}.

\begin{table*}[!htbp]
\centering
\caption{Comparison of AI responses with and without RAG to the question: \textit{"Which number should I call?"} with the context: \textit{"I'm alone."} and an image of fire.}
\label{tab:rag_call_number}
\begin{tabular}{@{\extracolsep{\fill}} p{3cm} p{13cm}}
\toprule
\textbf{Response Type} & \textbf{Response Content} \\
\midrule

\textbf{Without RAG} & 
If you're unsure about which number to call for assistance, please take a moment to gather your thoughts and consider the following steps: \newline
• \textbf{Stay calm}: Take a few deep breaths and try to relax. Panic can cloud your judgment, so it's essential to remain level-headed. \newline
• \textbf{Gather information}: Think about what you know about the situation that requires help. Make a mental note of any relevant details, such as the location or the type of assistance needed. \newline
• \textbf{Reach out for support}: If possible, contact a trusted friend, family member, or neighbour for guidance and support. Sometimes, talking through your concerns with someone who cares about you can provide clarity. \newline
• \textbf{Call 911 (or local emergency number)}: If you're still unsure about what number to call, don't hesitate to reach out to emergency services. They will assess the situation and provide guidance on the next steps. \newline
\textit{Remember, it's always better to err on the side of caution when seeking help. Take your time, gather your thoughts, and trust that you'll make the right decision for yourself and those around you.} \\
\midrule

\textbf{With RAG} & 
• Before you call, make sure you're in the right country to ensure you're contacting the correct emergency services. \newline
• Check your surroundings for any specific phone numbers or codes that might be relevant (e.g., 112 works across many European countries). \newline
• Consider using a translation app on your phone to help with language barriers if necessary. \newline
• If you're still unsure, try asking someone nearby for assistance; they may know the correct number to dial. \newline
\textit{Remember, safety is everyone's responsibility. Stay calm, take your time, and prioritise getting help when needed.} \\
\bottomrule
\end{tabular}
\end{table*}

As observed in \autoref{tab:rag_call_number}, the response without RAG provided generic emergency guidance and suggested calling 911 or another local emergency number without explicitly identifying the correct emergency number for Portugal. The RAG-enhanced response correctly identified the emergency number applicable in Portugal (112) but included advice for the user to verify their current country. This suggestion may inadvertently give the impression that the system is uncertain about the incident's location, despite clearly having access to location-specific data. Therefore, while the RAG implementation successfully provided accurate and relevant information, the prompt formulation could be further refined to avoid potential confusion regarding the system's knowledge about the user's geographical context.

\section{Discussion} \label{sec:discussion}

The evaluation of the Project Riley chatbot prototype confirmed the practical viability and effectiveness of the proposed architecture. The results validate that the multi-agent, emotion-informed framework performs reliably across various emotional and contextual domains. The system notably performed better in professional and family-oriented scenarios but showed comparatively weaker performance in emotionally complex situations such as breakups or interpersonal loss.

In the \textit{Emotional Appropriateness} category, the highest scores occurred in the themes of \textit{Job Loss or Unemployment} and \textit{Difficult Personal Decisions}, specifically for the question ``Did the final response seem emotionally appropriate to the context of your question?'', with both themes achieving a mean score of 4.71. This indicates the system's capability to effectively match emotional tone and content in structured or goal-focused scenarios. Conversely, responses concerning \textit{Breakups or Friendship Loss} consistently yielded the lowest mean scores in empathy (4.12) and appropriateness (4.41), highlighting challenges in capturing the complexity and variability inherent to interpersonal emotional dynamics.

This trend was further supported by the alignment between user-identified emotions and those suggested by the system, evaluated via the question ``Do you believe the answer(s) of the emotion(s) with the most votes were the most appropriate?''. The strongest alignment appeared in \textit{Family or Intergenerational Conflicts} (mean score: 4.35), indicating effective resonance with user expectations. In contrast, \textit{Breakups or Friendship Loss} scored the lowest (mean score: 3.94), suggesting a need for improved modelling of emotional subtleties in relational contexts.

Regarding \textit{Clarity and Utility}, results were generally positive. Clarity received high ratings across all themes, particularly for \textit{Job Loss or Unemployment}, with a maximum score of 4.71 for the question ``Was the final response clear and understandable?''. However, perceived utility demonstrated greater variation. The lowest utility scores were associated with \textit{Breakups or Friendship Loss} (3.88), whereas \textit{Difficult Personal Decisions} and \textit{Family or Intergenerational Conflicts} achieved higher scores (both 4.29). This underscores the importance of contextual relevance and insightful content in enhancing perceived utility.

Visualisation support was the least positively evaluated aspect, with scores ranging between 3.76 and 4.24. This suggests that, although conceptually beneficial, visualisations of emotion selection were insufficiently intuitive or impactful, warranting further interface refinement to effectively support user comprehension and reflection.

Lowest overall scores were recorded in the \textit{Naturalness and Humanisation} category. Even the highest rated theme, \textit{Family or Intergenerational Conflicts}, did not surpass a mean score of 3.88, highlighting significant limitations in achieving human-like conversational expression. Particularly, \textit{Anxiety in Academic or Professional Contexts} scored lowest (3.59), emphasising the necessity to enhance naturalness and linguistic subtlety in emotionally nuanced interactions.

Analysis of open-text responses regarding predominant emotions indicated that the system’s affective outputs generally aligned with user queries. For instance, Joy predominated in \textit{Anxiety in Academic or Professional Contexts}, reflecting a reassuring tone, while Fear was most prevalent in \textit{Job Loss or Unemployment} and \textit{Difficult Personal Decisions}, aligning well with anticipated emotional reactions. \textit{Breakups or Friendship Loss} was predominantly associated with Sadness and Disgust, reinforcing the emotionally complex nature inadequately addressed by the system.

Looking to Armando chatbot, the RAG evaluation showed clear advantages in content specificity and contextual accuracy. When responding to specific queries (e.g., ``Where is the fire happening?''), RAG-enhanced responses consistently outperformed non-RAG responses by providing accurate, contextually relevant, and actionable information. This demonstrates the crucial role of verified external knowledge integration in emergency communication scenarios.

However, the query ``Which number should I call?'' highlighted challenges in clearly communicating geographical specificity. Although the system correctly identified Portugal’s emergency number (112), suggesting that users confirm their location introduced unnecessary ambiguity regarding the chatbot’s geographical awareness. Future implementations should minimise such ambiguity by presenting clear and singular region-specific recommendations, avoiding conditional or uncertain expressions.

Consequently, enhancing the input quality for the RAG embedding process is essential. Establishing structured syntax and annotation protocols for input textual data could significantly improve the chatbot’s ability to prioritise and retrieve relevant contextual information. Clearly defined markers for urgency, location specificity, and emergency contacts would empower the chatbot to deliver precise and reliable guidance during critical events.

In conclusion, while the Armando RAG-enhanced chatbot prototype showed substantial promise in delivering timely and contextually appropriate responses during emergencies, further refinement in information retrieval granularity and response clarity is necessary to ensure optimal effectiveness and reliability in critical situations.

\section{Conclusion and Future Work}

This work presented, implemented, and tested Project Riley, a conversational AI architecture leveraging generative AI technologies. Inspired by Pixar’s 'Inside Out' films, the architecture delivers emotionally aware responses, adaptable to various psychological models of emotion. Additionally, this architecture was specifically adapted into a chatbot named Armando, designed to support citizens in post-disaster situations by providing emotionally calibrated responses combined with reliable and authoritative information to mitigate panic and anxiety.

Future research could explore the use of separately fine-tuned LLM, each specifically retrained to reflect a distinct emotional state. This approach would enable an empirical examination of whether such specialised models produce meaningfully different responses when compared to those generated by a shared model conditioned on emotion. Clearly defined objective metrics for assessing model performance are essential, along with comprehensive user testing to gather feedback on the contextual appropriateness and emotional resonance of the chatbot's responses.

Further improvements should prioritise enhancing user interaction by making the visualisation of the emotional reasoning process, comprising discussion, voting, and reasoning,optional or more discreet. In addition, prompt formulations could be adapted according to the type of situation addressed, allowing for more immersive and context-sensitive responses. Finally, in the context of emergency communication, it is crucial to establish a structured syntax and annotation protocol for RAG input data, to ensure more accurate retrieval and delivery of critical information.

\section{Competing interests}
No competing interest is declared.

\section{Author contributions statement}

ARO and DF conceived the study. NC and AP secured funding. ARO, GV, DF, and LF carried out the investigation. ARO, GV, DF, LF, NC, and AP developed the methodology. DF, LF, NC, and AP provided resources and supervised the work. ARO and DF developed the software. ARO, GV, DF, LF, NC, and AP validated the results. ARO, GV, DF, and LF drafted the original manuscript, and all authors revised and edited the final version.

\section{Data Availability}
The data underlying this article will be shared on reasonable request to the corresponding author.

\section{Funding}
This work was supported by FCT – Fundação para a Ciência e Tecnologia, I.P., under the project UIDB/04524/2020.

\section{Acknowledgments}

During the preparation of this work the authors used ChatGPT in order to improve the readability and language of the manuscript. After using this tool/service, the authors reviewed and edited the content as needed and take full responsibility for the content of the published article.

\bibliographystyle{unsrt}
\bibliography{reference}



\begin{biography}{{\includegraphics[width=77pt,height=77pt,clip,keepaspectratio]{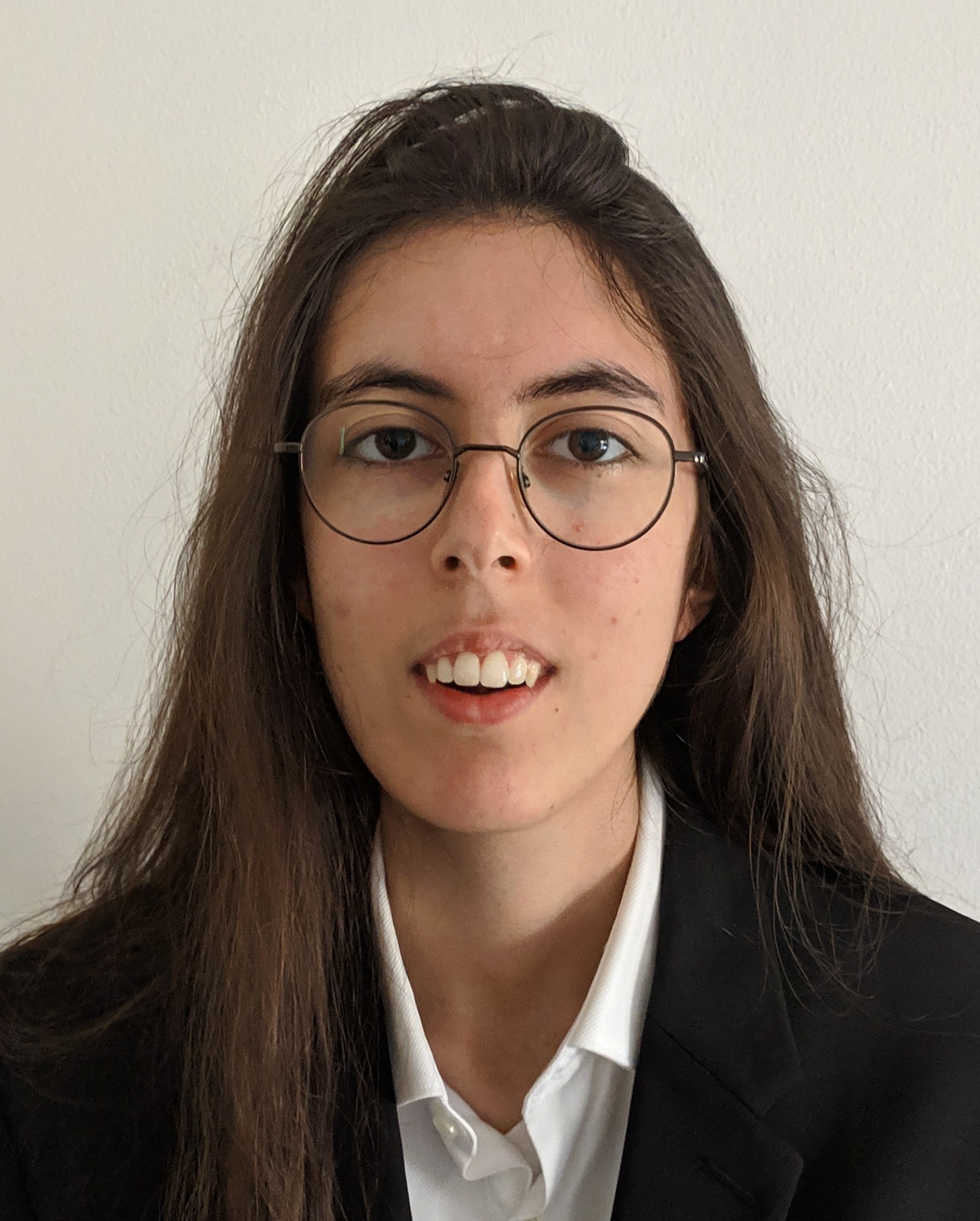}}}{\author{Ana Rita Ortigoso} was born in Portugal, in 2002. She received the B.Sc. degree in Computer Engineering from the Polytechnic University of Leiria, Leiria, Portugal, in 2023. She is currently pursuing the M.Sc. degree in Cybersecurity and Computer Forensics at the same institution.

Since 2023, she has been a research fellow at Computer Science and Communication Research Centre (CIIC), Polytechnic University of Leiria. Her research interests include wireless communication networks, the Internet of Things (IoT), Software-Defined Networking (SDN), emergency networks, and Large Language Models (LLMs).}
\end{biography}

\begin{biography}{{\includegraphics[width=77pt,height=77pt,clip,keepaspectratio]{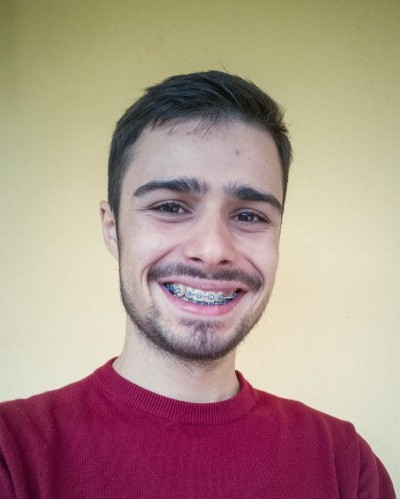}}}{\author{Gabriel Vieira} was born in Portugal in 1999. He received the B.Sc. degree in Computer Engineering, with a specialisation in Information Technology in Polytechnic University of Leiria, Portugal. He is currently pursuing, in the same school, the M.Sc. degree in Cybersecurity and Informatics. 

Since 2023, he has been a Research Fellow at the Computer Science and Research Centre, Polytechnic University of Leiria, Portugal. His research interests include cybersecurity, systems architecture, and emerging digital technologies. Mr. Vieira is passionate about technology, often engaging with new developments beyond academic or professional contexts.}
\end{biography}

\begin{biography}{{\includegraphics[width=77pt,height=77pt,clip,keepaspectratio]{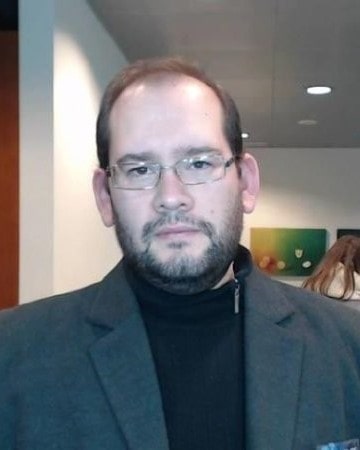}}}
{\author{Daniel Fuentes} was born in Venezuela, in 1984. He received the B.Sc. degree in Computer Engineering – Information and Communication Technologies in 2011 and the M.Sc. degree in Computer Engineering – Mobile Computing in 2013, both from the Polytechnic Institute of Leiria, Leiria, Portugal. He obtained the Ph.D. degree in Informatics from the University of Trás-os-Montes e Alto Douro, Vila Real, Portugal, in 2022. His doctoral research focused on “Software-Defined Things in Support of the Elderly.”

From 2011 to 2013 and from 2016 to 2022, he worked as an Invited Assistant at the Polytechnic Institute of Leiria. Since then he has been an Adjunct Professor in the Department of Computer Engineering at the School of Technology and Management (ESTG), Polytechnic Institute of Leiria. He is a Researcher at the Computer Science and Communication Research Centre (CIIC), where he is affiliated with the Smart IoT Ecosystems cluster. He is also the responsible of the Laboratory of Computer Applications (LAI12.DEI) and is member of the Scientific Council of CIIC. Additionally, he collaborates with the Cybersecurity and Computer Forensics Laboratory (LabCIF), contributing to digital forensic work in cooperation with the Public Prosecutor’s Office and the Judiciary Police.

His professional background includes experience as a researcher at INOV INESC Inovação – Institute of New Technologies, focusing on wireless communication networks and their configuration mechanisms. In terms of relevant professional experience in the industry, he was CEO and CTO of an IT startup, having developed several projects for clients in the areas of IoT, computer security and communication networks. His main interests lies in the IT field, namely in intelligent computer networks, cybersecurity, the Internet of Things (IoT), software-defined systems and Artificial Intelligence (AI) applied to security, IoT and communication networks.
}   
\end{biography}

\begin{biography}{{\includegraphics[width=77pt,height=77pt,clip,keepaspectratio]{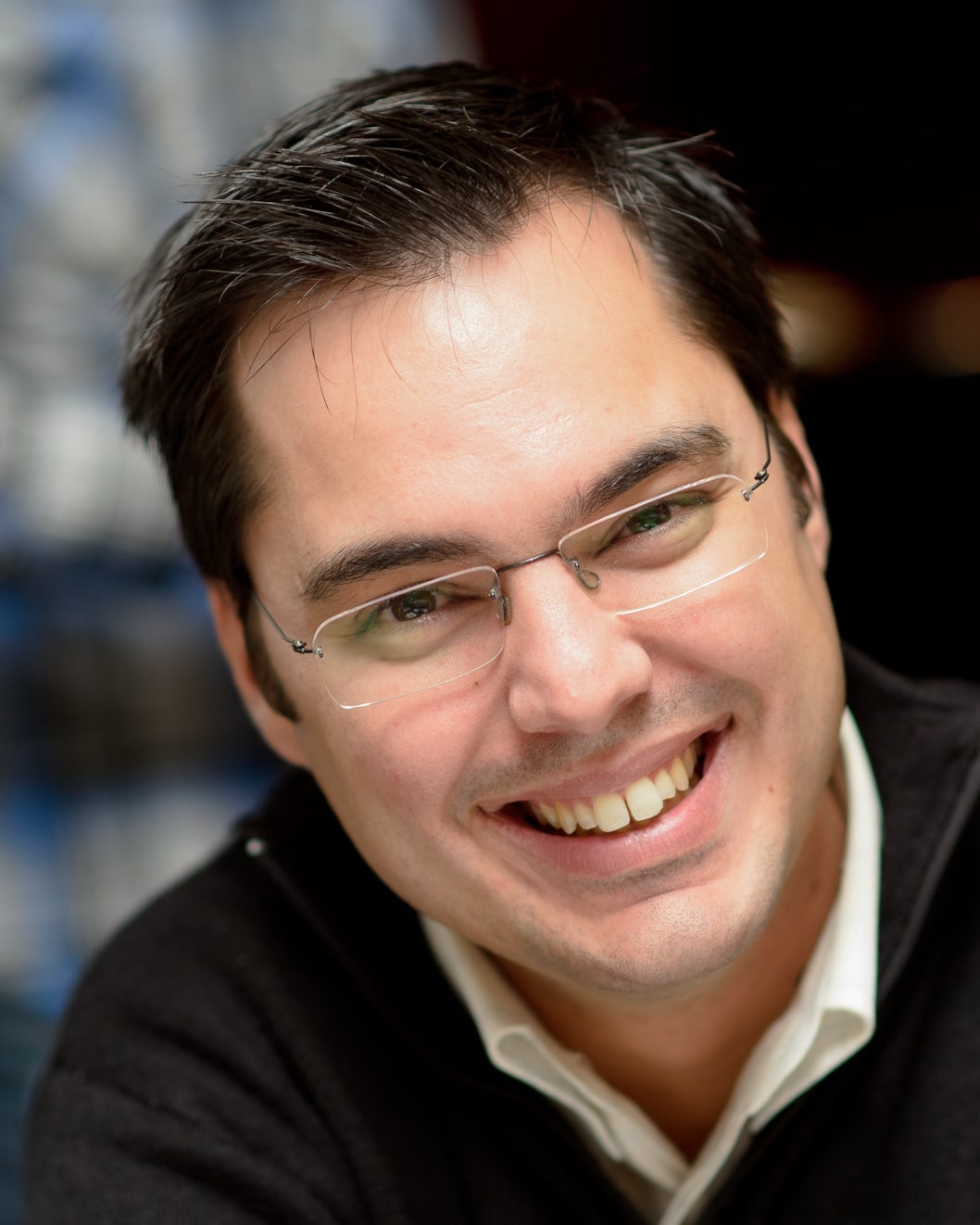}}}{
\author{Luís Frazão} was born in Portugal in 1983. He received the B.Sc. and M.Sc. degrees in Computer Engineering from the Polytechnic Institute of Leiria, Leiria, Portugal, in 2007 and 2011, respectively, and the Ph.D. degree in Intelligent and Adaptive Software Systems from the University of Vigo, Vigo, Spain, in 2017.

From 2007 to 2011, he worked as a network engineer at the Polytechnic Institute of Leiria. Since 2011, he has been with the School of Technology and Management (ESTG), Polytechnic Institute of Leiria, where he is currently an Adjunct Professor in the Department of Computer Engineering. He is a Senior Researcher at the Computer Science and Communication Research Centre (CIIC), where he serves as Deputy Director and is integrated in the Smart IoT Ecosystems cluster. He is also Head of the Communication Networks and Services Laboratory (2024–2026), and coordinates the Cisco Networking Academy and AWS Academy at ESTG. His teaching activities cover topics such as network and systems security, secure services, and internet technologies. He is a collaborator at the Cybersecurity and Computer Forensics Laboratory (LabCIF), contributing to digital forensic work in collaboration with the Public Prosecutor’s Office and the Judiciary Police.

His research interests include the Internet of Things (IoT), wireless communications, cybersecurity, network security, and network management and monitoring. He has participated in several national and international research projects and supervised multiple research fellows.}
\end{biography}

\begin{biography}{{\includegraphics[width=77pt,height=77pt,clip,keepaspectratio]{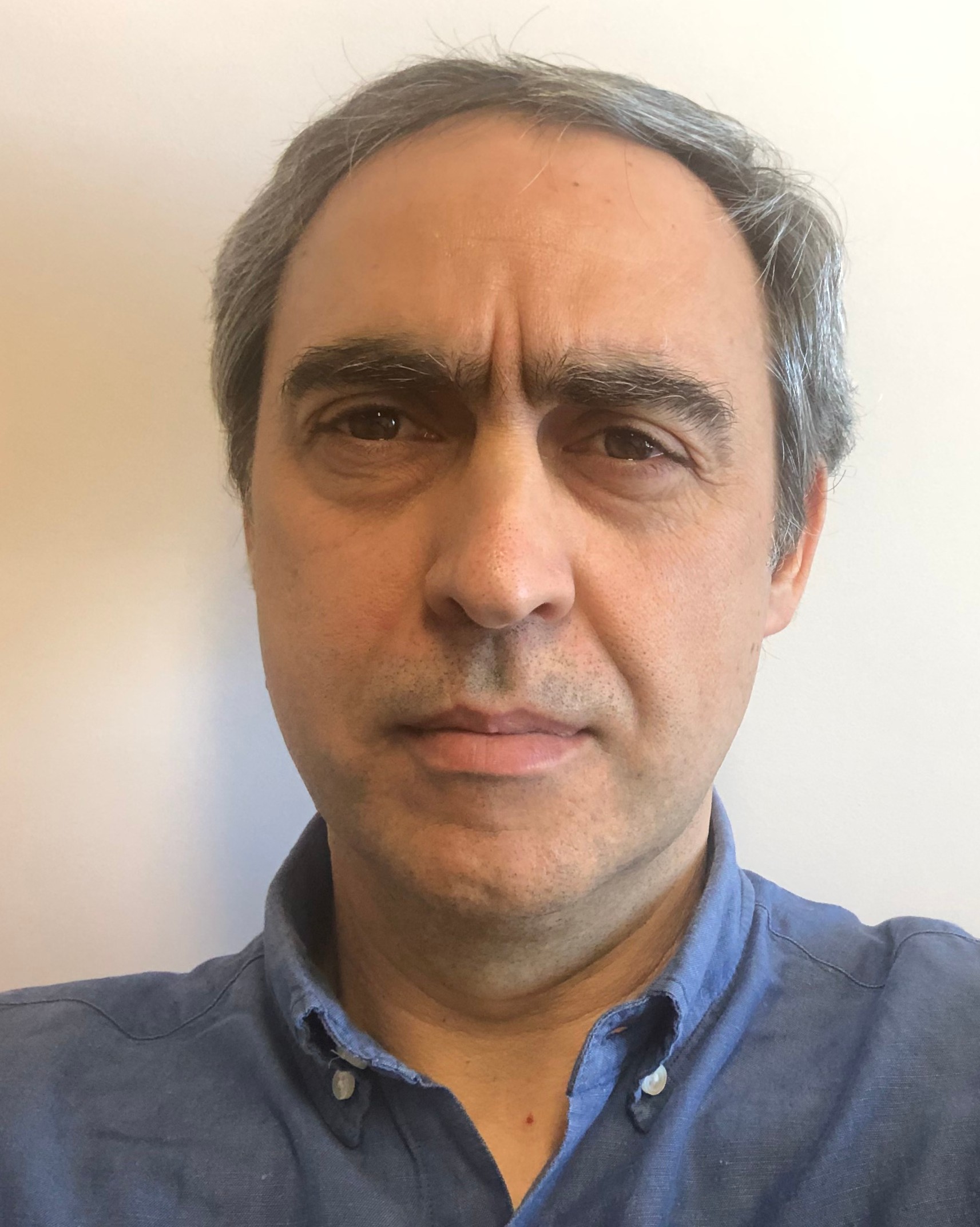}}}
{\author{Nuno Costa} was born in Portugal. He received the B.Eng. degree in Informatics and Systems Engineering from the Polytechnic Institute of Coimbra, Coimbra, Portugal, in 1995, the M.Sc. degree in Informatics Engineering from the University of Coimbra, Coimbra, Portugal, in 2006, and the Ph.D. degree in Informatics from the University of Trás-os-Montes e Alto Douro, Vila Real, Portugal, in 2010.

From 2001 to 2003, he worked at Critical Software, Portugal. He was a Research Intern at the Laboratory for Automation and Systems of the Pedro Nunes Institute, Coimbra, Portugal, in 2004. Since 2004, he has been with the School of Technology and Management (ESTG) at the Polytechnic Institute of Leiria, where he is currently a Coordinator Professor in the Department of Computer Science Engineering. He is also a member of the Computer Science and Communications Research Centre (CIIC) at the same institution. He has taught and coordinated courses in Systems Integration and Interoperability, Wireless Networks, and Platforms for IoT, among others. He has supervised numerous undergraduate and master's projects related to IoT, systems integration, and wireless communications.

His research interests include standardized and semantic interoperability, smart systems, wireless sensor networks, and embedded systems. He has authored and co-authored several papers in journals and conference proceedings in these areas. He is also active in reviewing articles for journals such as Sensors, Electronics, and Technologies (Hindawi), Sustainability (MDPI), and the Journal of Ambient Intelligence and Smart Environments.}
\end{biography}

\begin{biography}{{\includegraphics[width=77pt,height=77pt,clip,keepaspectratio]{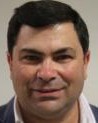}}}
{\author{António Pereira} was born in Portugal in 1967. He received the B.Sc. and M.Sc. degrees in Electronics and Telecommun\-ications Engineering from the University of Aveiro, Aveiro, Portugal, in 1992 and 1995, respectively, and the Ph.D. degree in Computer Science Engineering from the University of Coimbra, Coimbra, Portugal, in 2006. He obtained the Habilitation degree in Informatics from the University of Aveiro in 2018.

He is currently a Full Professor at the Polytechnic Institute of Leiria, Leiria, Portugal, where he is the head of the Smart IoT Ecosystems research line at the Computer Science and Communications Research Centre (CIIC). He has been a researcher at INOV INESC Inovação since 2008. He was the founder of CIIC and has held various leadership roles, including Coordinator of the Master in Computer Engineering – Mobile Computing and Head of the Computer Science Engineering Department. He has supervised 14 Ph.D. and 55 M.Sc. theses and is currently supervising 4 Ph.D. and 3 M.Sc. students in areas such as Industry 4.0, Internet of Things, Internet of Unmanned Vehicles, Cybersecurity, and Active Assisted Living.

Prof. Pereira has authored over 150 publications in conferences and refereed journals in the fields of Computer Science and Communications. He holds three national patents and has received two best paper awards at international conferences. He has participated in numerous Portuguese, Spanish, and European research projects and has coordinated more than 50 R\&D+I projects for small and medium-sized enterprises and institutions. His research interests include the Internet of Things, Smart IoT Ecosystems, Internet of Unmanned Vehicles, Ambient Assisted Living, Next Generation Networks and Services, Industry 4.0, and Large-Scale Intelligent Systems.}
    
\end{biography}

\end{document}